\documentclass{article}

\usepackage{arxiv}

\usepackage{forest}
\useforestlibrary{edges}
\usepackage{subcaption}
\usepackage{threeparttable}
\usepackage{amsmath}
\usepackage{lipsum}
\usepackage{multirow}
\usepackage{blindtext}
\usepackage{amssymb}
\usepackage{soul}
\usepackage{url}
\usepackage{geometry, color, graphicx, float}
\usepackage[colorlinks]{hyperref}
\usepackage{acronym}
\usepackage{natbib}
\usepackage{doi}
\usepackage{tikz}
\usepackage{amssymb}

\usepackage{amsmath,amssymb}
\usepackage{algorithm}
\usepackage[noend]{algpseudocode}
\usepackage{siunitx}
\usepackage{subcaption}
\usepackage{booktabs}
\usepackage{multirow}

\usepackage{array}      
\usepackage{booktabs}   
\usepackage{multirow}   
\usepackage{siunitx}    

\newcolumntype{L}[1]{>{\raggedright\arraybackslash}p{#1}}

\newcommand{\tblsetup}{%
  \small
  \setlength{\tabcolsep}{4pt}
  \renewcommand{\arraystretch}{1.1}
}

\usepackage{dblfloatfix}  

\title{STDiff: A State Transition Diffusion Framework for Time Series Imputation in Industrial Systems}

\author{ \href{https://orcid.org/0000-0000-0000-0000}{\includegraphics[scale=0.06]{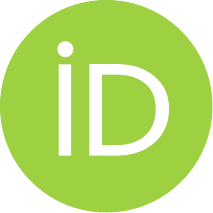}\hspace{1mm}Gary Simethy}\\
	Department of Energy Technology\\
	Aalborg University\\
	Esbjerg, Denmark 6700 \\
	\texttt{gasi@energy.aau.dk} \\
	\And
	\href{https://orcid.org/0000-0000-0000-0000}{\includegraphics[scale=0.06]{orcid.pdf}\hspace{1mm}Daniel Ortiz-Arroyo} \\
	Department of Energy Technology\\
	Aalborg University\\
	Esbjerg, Denmark 6700 \\
	\texttt{doa@energy.aau.dk} \\
	\And
	\href{https://orcid.org/0000-0000-0000-0000}{\includegraphics[scale=0.06]{orcid.pdf}\hspace{1mm}Petar Durdevic} \\
	Department of Energy Technology\\
	Aalborg University\\
	Esbjerg, Denmark 6700 \\
	\texttt{pdl@energy.aau.dk} \\
}

\date{}

\renewcommand{\shorttitle}{STDiff: A State Transition Diffusion Framework for Time Series Imputation}

\hypersetup{
pdftitle={A template for the arxiv style},
pdfsubject={q-bio.NC, q-bio.QM},
pdfauthor={David S.~Hippocampus, Elias D.~Striatum},
pdfkeywords={First keyword, Second keyword, More},
}

\begin{document}
\maketitle

\begin{abstract}
Incomplete sensor data is a major obstacle in industrial time-series analytics. In wastewater treatment plants (WWTPs), key sensors show long, irregular gaps caused by fouling, maintenance, and outages. We introduce STDiff and STDiff-W, diffusion-based imputers that cast gap filling as state-space simulation under partial observability, where targets, controls, and exogenous signals may all be intermittently missing. STDiff learns a one-step transition model conditioned on observed values and masks, while STDiff-W extends this with a context encoder that jointly inpaints contiguous blocks, combining long-range consistency with short-term detail. On two WWTP datasets (one with synthetic block gaps from Agtrup and another with natural outages from Avedøre), STDiff-W achieves state-of-the-art accuracy compared with strong neural baselines such as SAITS, BRITS, and CSDI. Beyond point-error metrics, its reconstructions preserve realistic dynamics including oscillations, spikes, and regime shifts, and they achieve top or tied-top downstream one-step forecasting performance compared with strong neural baselines, indicating that preserving dynamics does not come at the expense of predictive utility. Ablation studies that drop, shuffle, or add noise to control or exogenous inputs consistently degrade NH$_4$ and PO$_4$ performance, with the largest deterioration observed when exogenous signals are removed, showing that the model captures meaningful dependencies. We conclude with practical guidance for deployment: evaluate performance beyond MAE using task-oriented and visual checks, include exogenous drivers, and balance computational cost against robustness to structured outages.
\end{abstract}


\section{Introduction}
High-frequency industrial monitoring streams routinely suffer from incomplete observations due to sensor fouling, maintenance, calibration, and telemetry outages, which can bias analysis if not handled carefully \citep{khayati2020mindthegap,kazijevs2023jbi}. Wastewater treatment plants (WWTPs) are a prominent example where irregular and often block-structured gaps are common, and where physically consistent reconstructions are required for safe decision support \citep{zhi2024deeplearningforwaterquality}.

Classical approaches span interpolation and smoothing, state-space and expectation-maximization (EM) methods, matrix/tensor completion, and motif-based pattern matching. EM-based imputers iteratively estimate missing values by alternating between estimating latent states and updating model parameters, typically assuming linear-Gaussian dynamics. These methods can work well under short or random missingness, but accuracy degrades when outages remove contiguous blocks and when exogenous drivers modulate the dynamics in nonstationary ways \citep{khayati2020mindthegap}. Recent evaluations and reviews emphasize that contiguous ``blackout'' gaps are especially challenging and that preserving trajectory shape matters beyond pointwise error \citep{khayati2020mindthegap,wang2024btaskimputation}.

Deep learning has advanced time-series imputation by learning directly from partially observed sequences with explicit masks and time gaps. Representative RNN-based methods include GRU-D, which injects masking and elapsed-time decays into the hidden state \citep{che2018grud}, and BRITS, which treats missing values as learnable variables in a bidirectional recurrent graph \citep{cao2018brits}. Transformer-style models improve long-range dependency capture; for example, SAITS reports strong results on healthcare and environmental data \citep{du2023saits}. Probabilistic diffusion models provide flexible conditional distributions for masked positions; CSDI is a widely used baseline that conditions a score-based model on observed entries \citep{tashiro2021csdi}. Beyond these imputation-centric lines, complementary families model continuous-time dynamics or emphasize multiscale forecasting: Latent ODE and GRU-ODE-Bayes accommodate irregular sampling via neural ODEs \citep{rubanova2019latentode,debrouwer2019gruodebayes}, and TimeMixer offers a decomposable multiscale mixing architecture for forecasting that has been adapted in practice for related tasks such as anomaly detection and imputation \citep{wang2024timemixer}.

\textbf{Two practical gaps} remain critical for industrial monitoring. \textit{First}, a large fraction of published evaluations assume missing-completely-at-random (MCAR) sampling, which can overstate accuracy relative to operational outages that remove contiguous blocks \citep{khayati2020mindthegap}. \textit{Second}, the system is \emph{partially observed} in practice: not only the target measurements but also control and exogenous inputs are intermittently missing, so models must condition on masked covariates yet still reconstruct trajectories that respect process physics \citep{zhi2024deeplearningforwaterquality}.

\subsection{Our approach} 
We propose \textsc{STDiff} and \textsc{STDiff-W} to address these challenges. Unlike earlier diffusion-based imputers such as CSDI, which condition only on observed entries and binary masks, \textsc{STDiff} incorporates control actions and exogenous signals to guide reconstructions toward physically plausible system dynamics. Similarly, while Latent ODE and other continuous-time models can model irregular sampling via neural ODEs, they do not simulate state transitions under partial observability. Our approach reframes imputation as a probabilistic rollout of system evolution, enabling robust recovery under long gaps and operational blackouts.

\textsc{STDiff} generates missing points step by step, effectively rolling out the time series as one would simulate a dynamical system. At each step, it conditions on the observed recent history, including the last observed state and any available control or exogenous inputs, to produce a probabilistic estimate of the next state \citep{che2018grud,cao2018brits}. By iterating, \textsc{STDiff} handles arbitrarily long gaps because it does not depend on a fixed-size observation window - an advantage for WWTPs where system dynamics and control actions drive the evolution.

Building on \textsc{STDiff}, \textsc{STDiff-W} integrates a windowed context encoder for multi-step imputation. The goal is to capture the best of both worlds: sequential rollouts often ensure long-range consistency, while window-based models can achieve lower immediate error on short gaps by leveraging recent patterns \citep{du2023saits,cao2018brits}. \textsc{STDiff-W} conditions the diffusion model on a recent-history window of length $K$ and directly imputes a block of $H$ consecutive missing steps in one pass, learning the joint distribution of a block conditioned on context. This reduces error accumulation within a gap and improves internal self-consistency of the filled segment. At inference time, \textsc{STDiff-W} slides its context window across the series and imputes any detected gap similarly to windowed baselines; stitching successive blocks enables gaps longer than $K$ to be imputed via a multi-stage strategy. The diffusion denoising backbone is shared with \textsc{STDiff}, but the guided context vector emphasizes recent trends.

Both variants incorporate control and exogenous signals as conditioning information. Signals are encoded alongside the state at each time step so that the model can learn how control actions influence transitions. Prior evaluations note that generic imputers tend to oversmooth and degrade the fidelity of spikes and regime changes in blackout settings, motivating explicit conditioning on drivers rather than purely data-driven interpolation \citep{khayati2020mindthegap,zhi2024deeplearningforwaterquality}. In our experiments, causal-style ablations that drop, shuffle, or noise controls and exogenous inputs consistently degrade accuracy, indicating that the learned dependencies are meaningful.

We evaluate on a real WWTP dataset with nonstationary behavior under high missing rates. We introduce controlled levels of synthetic block gaps and benchmark against strong RNN, Transformer, diffusion, continuous-time, and multiscale baselines, including BRITS \citep{cao2018brits}, GRU-D \citep{che2018grud}, SAITS \citep{du2023saits}, CSDI \citep{tashiro2021csdi}, Latent ODE \citep{rubanova2019latentode}, and TimeMixer \citep{wang2024timemixer}. Beyond point-error metrics, we evaluate task-oriented utility by training a simple one-step forecaster on each method’s imputed training data and evaluating it on that method’s imputed test data. \textsc{STDiff-W} maintains strong downstream predictive performance even when not the lowest on imputation MAE, while some lower-MAE baselines underperform on forecasts - reinforcing the value of task-oriented evaluation \citep{wang2024btaskimputation}.

\subsection{Contributions}
\begin{enumerate}
    \item A diffusion-based imputation framework that treats reconstruction as state-space simulation under partial observability.
    \item An enhanced windowed variant (\textsc{STDiff-W}) that combines sequential simulation with blockwise diffusion conditioned on causal context windows - a novel hybrid approach that mitigates error accumulation in long rollouts and is, to our knowledge, absent in prior imputation models.
    \item An evaluation on WWTP data with blocky synthetic gaps aligned to operations and a dataset with natural outages, plus comparisons to RNN, Transformer, diffusion, continuous-time, and multiscale baselines.
    \item Causal-style ablations showing that conditioning on controls and exogenous signals is critical for fidelity.
    \item A downstream forecasting study showing that preserving dynamics and variance can matter more than pointwise MAE for operational utility.
\end{enumerate}

The remainder of this paper is organized as follows: Section~2 reviews related work on time-series imputation. Section~3 describes the proposed methodology, including the STDiff model and the STDiff-W extension. Section~4 details the experimental setup and datasets. Section~5 presents the results of our quantitative and qualitative evaluations. Finally, Sections~6 and~7 discuss the findings and conclude the paper.

\section{Related Work}
This section reviews prior work on time-series imputation, covering both classical techniques and modern deep learning approaches, as well as evaluation considerations under realistic outages.

\subsection{Data quality and missing data in industrial systems}

The need to handle missing or corrupted sensor data is well established in industrial analytics. Early work rooted in state-space modeling and EM/Kalman smoothing treated missing observations within probabilistic filtering/smoothing pipelines, laying a foundation for principled reconstruction under linear-Gaussian assumptions \citep{shumway1982em}. Large-scale evaluations show that outages often appear as contiguous blocks rather than isolated points and that method rankings can change drastically under blackout masking, motivating metrics beyond pointwise error and a focus on trajectory/physical consistency \citep{khayati2020mindthegap}.

\subsection{Traditional imputation methods}

Simple heuristics - mean/median fills, \textbf{Last Observation Carried Forward (LOCF)} / \textbf{Next Observation Carried Backward (NOCB)}, and linear/spline interpolation remain common baselines but implicitly assume no new dynamics within gaps. Tooling such as \textbf{imputeTS} documents practical variants and their assumptions for time series \citep{moritz2025imputets}. While splines capture smooth nonlinearity, they can over-smooth spikes unless constrained \citep{bilos2022splinenet,moritz2025imputets}. 

Classical time-series models such as \textbf{ARIMA}/SARIMA and \textbf{state-space/Kalman} filters have also been used for imputation. ARIMA-type models assume (seasonal) stationarity and rely on contiguous samples to construct lagged terms and error corrections; gaps fundamentally disrupt these lagged structures and make estimation unstable \citep{magyari2025riverflow}. Recent evaluations on hydrological and other environmental series report that seasonal ARIMA “struggles with consecutive missing values’’ and rapidly loses accuracy as missing segments lengthen \citep{farjallah2025missingts}. Under model-based assumptions, \textbf{Expectation-Maximization (EM)} with Kalman smoothing offers strong solutions when a well-specified linear-Gaussian state-space model and frequent observations are available \citep{shumway1982em}, but theoretical results on Kalman filtering with intermittent observations show that below a minimum observation rate the error covariance diverges \citep{sinopoli2004kalmanintermittent}. In practice, these methods degrade sharply under long, structured outages and strongly exogenous-driven regimes, as also observed in recent comparative studies \citep{magyari2025riverflow,farjallah2025missingts}.

\textbf{Multiple Imputation by Chained Equations (MICE)} is widely used when data are \textbf{Missing at Random (MAR)} \citep{vanbuuren2011mice}. Instance-based \textbf{k-Nearest Neighbors (KNN)} imputation leverages local similarity but degrades when close analogues are absent, a frequent failure mode under long blackouts or nonperiodic regimes \citep{batista2002knnimpute}. Because classical ARIMA/SARIMA and Kalman-based imputers are known to break down under the long, blackout-style gaps considered in this work, we treat them as part of the methodological background rather than as main baselines in our empirical comparisons.

\subsection{RNN-based imputation}

Recurrent models were early to encode masks and elapsed times explicitly. \textbf{GRU-D} down-weights stale features via time-gap decays \citep{che2016grud,che2018grud}. \textbf{BRITS} treats missing values as variables optimized via bidirectional consistency and remains a strong baseline \citep{cao2018brits}. Non-recurrent formulations (e.g., \textbf{NRTSI}) address irregular sampling without recurrence \citep{shan2021nrtsi}. Reviews consistently note over-smoothing on long gaps and sensitivity to high missingness, echoing blackout cautions in \citep{khayati2020mindthegap}.

\subsection{Transformer and attention models}

Transformer-style imputers improve long-range dependence capture. \textbf{SAITS} introduces diagonally masked self-attention and reports strong results on healthcare and environmental benchmarks \citep{du2023saits}. Cross-dimensional attention (e.g., \textbf{CDSA}) exploits correlations across features, time, and locations \citep{ma2019cdsa}. For forecasting architectures adapted to imputation, \textbf{iTransformer} inverts tokenization to attend across variables \citep{liu2023itransformer}. In practice, many Transformer pipelines rely on fixed windows for tractability, limiting context unless paired with hierarchical encoders or recurrence \citep{du2023saits,liu2023itransformer}.

\subsection{CNNs and other forecasting backbones used for imputation}

Convolutional backbones efficiently model local structure and are often adapted for gap filling via reconstruction losses under random masks. Examples include \textbf{SCINet} (sample convolution and interaction) \citep{liu2021scinet}, \textbf{TimesNet} (2D temporal-variation modeling) \citep{wu2023timesnet}, and strong linear decompositions such as \textbf{DLinear}, which can rival large Transformers in long-horizon forecasting - hence their use as priors in hybrid imputers \citep{zeng2023dlinear}. Feature-wise modulation (\textbf{Temporal FiLM}) extends CNN receptive fields for long sequences \citep{birnbaum2019tfilm}.

\subsection{Graph neural networks for relational sensors}

When sensors are spatially related, graph inductive biases help. \textbf{GRIN} performs bidirectional message passing and reports sizable gains on traffic and environmental datasets \citep{cini2021grin}. Spectral approaches such as \textbf{StemGNN} apply a \textbf{Discrete Fourier Transform (DFT)} in time and a graph Fourier transform across series, influencing later spatiotemporal imputers and forecasters \citep{cao2020stemgnn}.

\subsection{Deep generative models: VAEs and GANs}

GP-VAE introduces a \textbf{Gaussian Process Variational Autoencoder}, which imposes a Gaussian Process (GP) prior in latent space to model smooth temporal evolution with uncertainty \citep{fortuin2020gpvae}. Generative Adversarial Network (GAN) - based imputers frame missing-value filling as a data generation problem. \textbf{GAIN (Generative Adversarial Imputation Networks)} popularized this setting and has shown competitive accuracy compared to MICE and missForest on clinical datasets \citep{yoon2018gain,dong2021gainclinical}. End-to-end GAN architectures (e.g., \textbf{E2GAN}) extend this approach to multivariate time series \citep{luo2019e2gan}. GANs can capture multimodal distributions but often require careful training stabilization.

\subsection{Continuous-time models}

Continuous-time latent dynamics enable modeling under irregular sampling and partial observability. \textbf{Latent Ordinary Differential Equation (Latent ODE)} and \textbf{ODE-RNN} hybrids generalize sequence models by introducing neural ODE dynamics in latent space \citep{rubanova2019latentode}. \textbf{GRU-ODE-Bayes} further couples continuous-time evolution with Bayesian measurement updates to handle uncertainty in irregular observations \citep{debrouwer2019gruodebayes}. \textbf{Neural Controlled Differential Equations (Neural CDEs)} ground the formulation in controlled differential equations, offering memory-efficient training and principled interpolation \citep{kidger2020ncde}. These methods serve as strong baselines for time series with irregular sampling and missingness.

\subsection{Diffusion, flow, and bridge models for time-series imputation}
\label{sec:rw-diffusion}

Mask-conditioned score models have been the dominant diffusion approach for imputation, with CSDI reporting early strong results \citep{tashiro2021csdi}. Subsequent developments add sequential inductive bias or efficiency (e.g., state-space backbones, observed-value consistency, spatiotemporal priors, intra-/inter-window consistency) \citep{lopez2022sssd,wang2023midm,liu2023pristi,zhou2024mtsci}. Recent surveys and sequence-focused works discuss conditioning for irregular and multivariate series, latent-space diffusion for multivariate imputation, function-space perspectives, and forecasting with score models \citep{yang2024tsdiffusionsurvey,meijer2024forecastingsurvey,liang2024lssdm,lim2025functionsdd,lim2024tfscore}.

To stabilize very long horizons, diffusion models have adopted windowed or autoregressive chunking with causal context and overlap; adjacent windows share information to mitigate error accumulation (long-horizon video and sequence generation provide recent formalizations) \citep{sun2025ardiffusion,gao2025ca2vdm}. The core idea is to denoise blocks jointly while conditioning on recent history, then slide the window to cover extended spans.

Probability-flow ODEs, conditional flows, and Schrödinger bridges offer faster sampling for imputation while retaining stochasticity \citep{chen2023csbi,qian2024clwf}. In parallel, physics-/constraints-informed diffusion introduces domain equations or penalties for (spatio-)temporal problems and neural SDE hybrids outline ways to encode dynamics \citep{wu2024piml,shu2023pidm,soni2025physdiff}.

These families differ mainly in (i) how conditioning is formulated (mask/window vs.\ additional structure such as controls/priors) and (ii) how long gaps are handled (one-step rollout vs.\ blockwise/windowed denoising). We articulate our research gap and design goals in Sec.~\ref{sec:gap}.

\subsection{Evaluation under realistic outages and downstream utility}

Evaluations under MCAR masking can overstate performance relative to industrial outages; \emph{blackout-style} gaps are essential for realism \citep{khayati2020mindthegap}. Recent work argues for \emph{task-oriented} evaluation of time-series imputation - for example, assessing downstream forecasting accuracy on imputed inputs rather than relying solely on point-error metrics. In particular, \citep{wang2024downstream} formalize forecasting-aware imputation evaluation, while \citep{kazijevs2023healthsurvey} highlight that no single method dominates across data types and missing patterns, emphasizing the need to assess imputation in context.

\subsection{Windowed versus full-sequence strategies}

A key design decision in time-series imputation is whether to train on sliding windows or to model full sequences. 
Windowed training increases sample count and batching efficiency, making optimization more stable, but inherently limits long-range temporal context. 
In contrast, \textbf{full-sequence methods} (e.g., \textbf{BRITS} or step-by-step state-space rollouts) can capture dependencies across arbitrarily long gaps but often face optimization and memory constraints \citep{cao2018brits}. 
Hybrid or two-stage designs combine local window reconstruction with global coherence - such as \textbf{RATAI}, which performs predictive imputation followed by reconstruction \citep{lai2024ratai}.  
Recent diffusion-based variants further enhance inter-window consistency or propagate long-range information through structured backbones \citep{zhou2024mtsci,lopez2022sssd}. 
These developments motivate architectures like \textsc{STDiff-W}, which conditions on a recent window for block inpainting while using sequential rollouts to handle extended gaps.

\subsection{Research gap and design goals}
\label{sec:gap}

Despite advances in time-series imputation, several limitations persist - particularly for industrial telemetry with structured outages, partial observability, and control signals. We summarize four key gaps:

\textbf{(G1) Robustness to structured outages.}
Many methods are benchmarked under MCAR/random masking, yet industrial systems often exhibit long, contiguous gaps with missing covariates. Models that perform well under random masks may over-smooth or drift during block missingness.

\textbf{(G2) Conditioning beyond masks.}
Diffusion models like CSDI condition only on observed values and binary masks, while physics-informed variants require full system equations. In practice, control actions and exogenous drivers are logged and informative, but physical models are often incomplete. This motivates conditioning that leverages available control context to improve realism.

\textbf{(G3) Long-gap error accumulation.}
One-step sequential rollouts maintain global continuity but can accumulate error over long gaps. Conversely, windowed denoising reduces local error but rarely targets multivariate transitions with control signals. There is a gap for hybrid models that combine both strategies.

\textbf{(G4) Task-oriented evaluation.}
Low pointwise error may obscure poor dynamical fidelity. Practical utility should be assessed via downstream tasks (e.g., forecasting) and visual plausibility on real outages.

\smallskip
To our knowledge, no existing imputation model integrates state-transition simulation with blockwise diffusion conditioned on causal context. This motivates our \textsc{STDiff-W} design, which bridges sequential simulation and joint inpainting to improve robustness under partial observability and long outages.

Guided by (G1)-(G4), we next formalize the imputation problem and introduce two complementary models: a one-step transition imputer (Sec ~\ref{sec:stdiff_methodolgy}) and a blockwise variant for long gaps (Sec ~\ref{sec:stdiffw_methodology}), followed by training and inference details.

\section{Methodology}

In this section, we formalize the imputation problem and present our proposed framework. Section~3.1 defines the problem setup. Section~3.2 describes the base one-step diffusion model (STDiff), and Section~3.3 introduces STDiff-W, which extends the approach with a context window for multi-step imputation. We also detail the diffusion process and masked inpainting strategy (Section~3.4) and the training/implementation specifics (Section~3.5).

\subsection{Problem formulation}
We consider a multivariate time series $\mathbf{X} = \{x_t\}_{t=1}^T$ with $x_t \in \mathbb{R}^d$. Let $\Omega \subset \{1, \dots, T\}$ be the set of observed time indices and $\bar{\Omega}$ the missing ones. The task is to estimate $\{x_t : t \in \bar{\Omega}\}$ using the statistical structure of $\mathbf{X}$ and any available controls or exogenous covariates $u_t \in \mathbb{R}^p$. In industrial plants, $u_t$ often includes setpoints and actuator signals that may be observed even when some sensors fail; thus both targets and covariates can be partially observed.

We model a one-step transition distribution $p(x_t \mid x_{t-1}, u_t)$. If $x_t$ is missing, we draw it from this conditional given the last available state and contemporaneous inputs. To represent flexible, possibly multi-modal transitions, we use a denoising diffusion probabilistic model (DDPM) trained with the standard noise-prediction objective \citep{ho2020ddpm}. 

For each observed pair $(x_{t-1}, x_t)$, we corrupt $x_t$ via a forward noising process and train a U-Net denoiser $\epsilon_\theta$ to predict the added Gaussian noise at diffusion step $k$, conditioned on a context vector $c_t = f([x_{t-1};\, u_t])$:
\[ \epsilon_\theta(x_t^{(k)}, k, c_t) \approx \epsilon, \]
where $x_t^{(k)}$ is the noised version of $x_t$ at step $k$ and $\epsilon \sim \mathcal{N}(0, I)$. This loss trains the model to infer the true next-state $x_t$ from noisy versions given the prior state and inputs.

At inference, we impute missing values via reverse-diffusion sampling from the learned diffusion model. Starting from a standard normal guess for the next state, we iteratively denoise:
\[ \hat{x}_{t_0+1} \sim p_\theta(\cdot \mid x_{t_0}, u_{t_0+1}), \quad \hat{x}_{t_0+2} \sim p_\theta(\cdot \mid \hat{x}_{t_0+1}, u_{t_0+2}), \dots, \] 
filling the gap sequentially. This yields simulator-like, unbounded-horizon imputation that incorporates control inputs at each step and, by sampling multiple trajectories, provides a natural way to quantify uncertainty.

\begin{figure}
    \centering
    \includegraphics[width=0.6\linewidth]{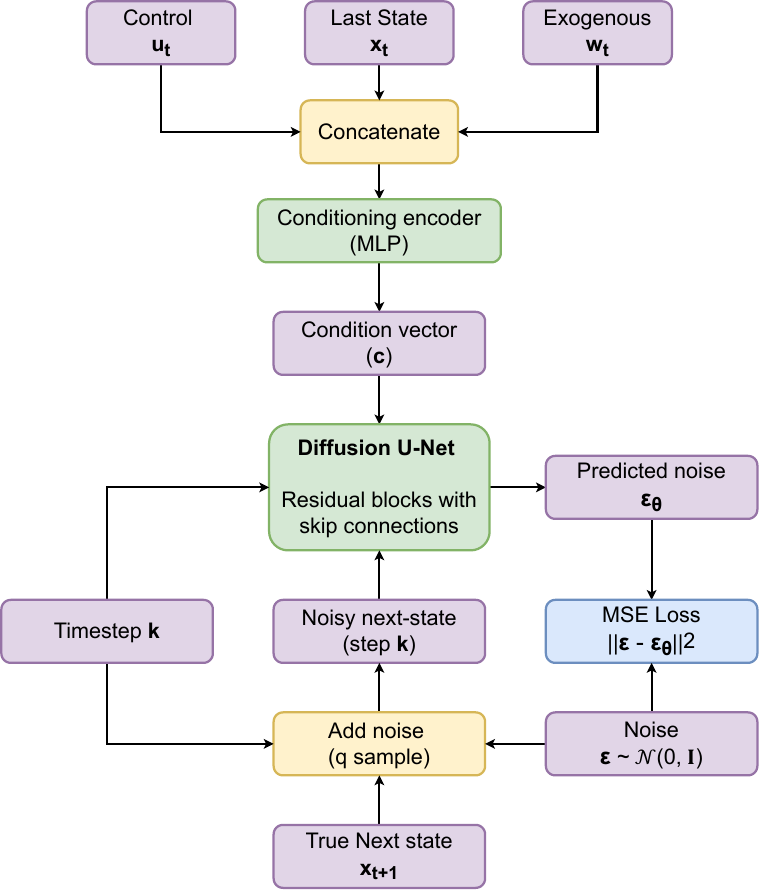}
    \caption{Architecture of \textbf{STDiff} during training. A 1D U-Net denoiser receives a noised version of the next state $x_t^{(k)}$ and predicts the added Gaussian noise, conditioned on the diffusion step $k$ and a context embedding of the previous state, control inputs, and exogenous variables. This trains a one-step transition model $p(x_t \mid x_{t-1}, u_t, w_t)$ that can later be used for sequential imputation.}
    \label{fig:stdiff_arch}
\end{figure}

\subsection{STDiff: one-step diffusion model for sequential imputation}
\label{sec:stdiff_methodolgy}

STDiff learns a single-step conditional diffusion model that generates the next state given the last observed state and current inputs. It imputes through a gap by rolling out one step at a time, akin to simulating a dynamical system. This sequential approach enables arbitrarily long gaps to be filled without a fixed window size.

\subsubsection{Forward and reverse diffusion}
We adopt a DDPM framework: the forward process adds Gaussian noise to the true next state $x_{t}$ over $T$ steps, and the reverse process progressively removes noise using a U-Net denoiser $\epsilon_\theta$. At each reverse step $k$, the model predicts the noise $\epsilon$ added at that step. The reverse kernel is defined by the standard DDPM parameterization conditioned on $\epsilon_\theta(x_t^{(k)}, k, c_t)$. The denoiser $\epsilon_\theta(\cdot)$ is conditioned on the context $c_t = g([x_{t-1};\, u_t])$, which is computed by a small embedding network (e.g., an MLP) encoding the previous state $x_{t-1}$ and any available inputs $u_t$.

\begin{figure}
    \centering
    \includegraphics[width=0.6\linewidth]{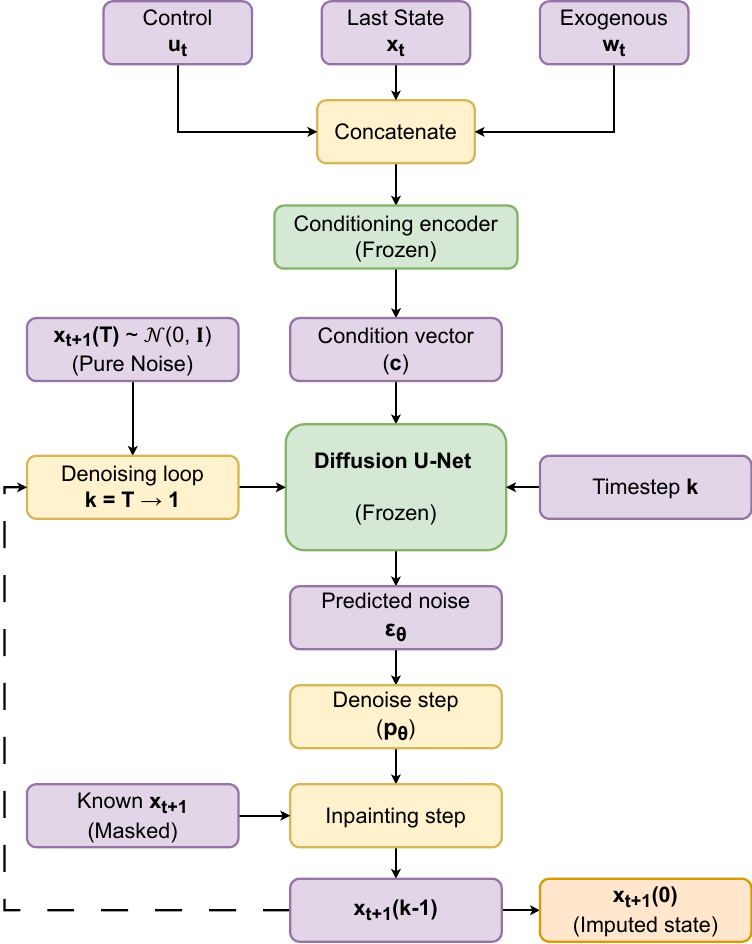}
    \caption{Imputation with \textbf{STDiff} at inference time. Starting from a Gaussian noise sample for the next state, the model iteratively denoises over the diffusion steps ($k = T \rightarrow 1$) while clamping any observed entries, yielding an imputed next state that respects available measurements. Repeating this procedure step by step through a gap produces a sequential rollout that fills arbitrarily long outages.}
    \label{fig:stdiff_infer}
\end{figure}

\subsubsection{Masked inpainting under partial observability}
Let $M \in \{0,1\}^{\text{dim}(x)}$ be a binary mask for observed entries (1 means observed). During each reverse step $k$, we clamp observed entries of the state and only update the unknown components. This ensures that if some dimensions of $x_t$ (or some sensors) are still operational at time $t$, their values remain unchanged during imputation. This masked inpainting rule keeps observed entries fixed while sampling the unknown portions (this process is depicted in Figure~\ref{fig:stdiff_infer}). 

\subsubsection{Architecture}
The denoiser is a 1D U-Net with residual blocks and skip connections, conditioned on the context $c_t$ and diffusion-step embeddings. The network predicts the Gaussian noise term $\epsilon$ added to the true next state, given the current diffusion timestep and the embedding of the previous state $x_{t-1}$, current control $u_t$, and current exogenous variables $w_t$. Figure~\ref{fig:stdiff_arch} illustrates the STDiff model architecture during training.

\subsubsection{Sequential imputation algorithm}
For a gap $(t_0+1, \dots, t_0+H)$, STDiff imputes sequentially as:
\[ \hat{x}_{t_0+1} \sim p_\theta(\cdot \mid x_{t_0}, u_{t_0+1}), \]
\[ \hat{x}_{t_0+2} \sim p_\theta(\cdot \mid \hat{x}_{t_0+1}, u_{t_0+2}), \]
and so forth, until $\hat{x}_{t_0+H}$. Each $\hat{x}$ is generated via the diffusion sampling (denoising) process with masked inpainting for any known entries. This procedure yields an imputed trajectory through the gap. By conditioning on the immediate past state and inputs at each step, STDiff naturally incorporates control actions and avoids a fixed receptive field. However, because each step only has access to the most recent history, errors can accumulate over very long gaps.

\textit{In summary, STDiff can roll out imputations over arbitrarily long horizons by iterative simulation. However, purely sequential generation may accumulate error over medium to long gaps. To mitigate this, we introduce STDiff-W, which leverages a context window for block-wise imputation.}

\subsection{STDiff-W: multi-step imputation with a context window}
\label{sec:stdiffw_methodology}

Recursive one-step rollouts can maintain long-range consistency, but they tend to
accumulate error over extended gaps. In contrast, \emph{direct multi-step} prediction mitigates this accumulation by
predicting several future points in one shot, which often improves accuracy in
forecasting and imputation \citep{taieb2012rectify}.
 Our idea is to provide the model with a broader temporal context and allow it to generate a block of future values at once. STDiff-W augments STDiff with a windowed context encoder and a block generator that imputes $H$ consecutive steps jointly. 

\begin{figure}
    \centering
    \includegraphics[width=0.6\linewidth]{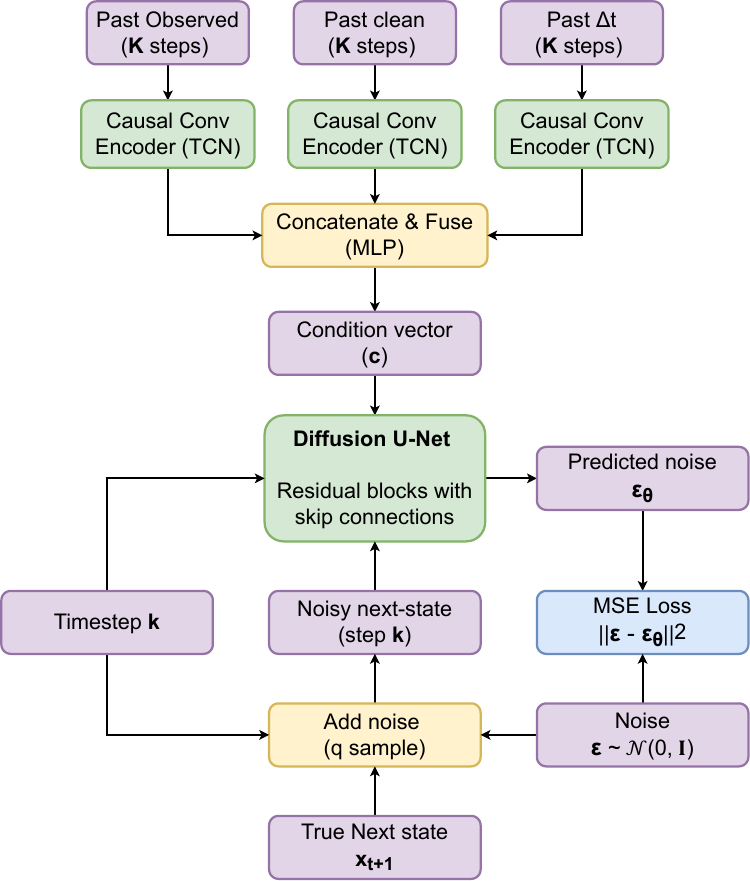}
    \caption{Architecture of \textbf{STDiff-W} during training. A causal temporal convolutional network (TCN) encodes the past $K$ steps of states, masks, and (optionally) $\Delta t$ into a context vector that summarizes recent dynamics. This context conditions a diffusion U-Net that jointly denoises a block of $H$ future steps, learning a blockwise transition model that mitigates error accumulation compared with pure one-step rollouts.}
    \label{fig:stdiffw_arch}
\end{figure}

\subsubsection{Design}
During training, we simulate random contiguous gaps of length $H$ and provide the most recent $K$ observed steps and covariates $\{(x_{t-K}, u_{t-K+1}), \dots, \allowbreak (x_{t-1}, u_{t})\}$ to a causal dilated temporal convolution (TCN) encoder, producing a context vector $c$ \citep{bai2018tcn}. The diffusion U-Net then denoises a $H \times d$ stacked target 
($H$ time steps, $d$ feature dimensions) in one pass, conditioned on $c$. This uses the same $\epsilon$-prediction objective as STDiff but applied over the joint $H$-step block. Causal TCNs offer long effective receptive fields while preserving order \citep{bai2018tcn}. The training and inpainting process for STDiff-W is analogous to STDiff but applied to blocks (Figure~\ref{fig:stdiffw_arch}). The loss is the mean squared error between the predicted noise and true noise across all $H$ time steps in the block.

\begin{figure}
    \centering
    \includegraphics[width=0.6\linewidth]{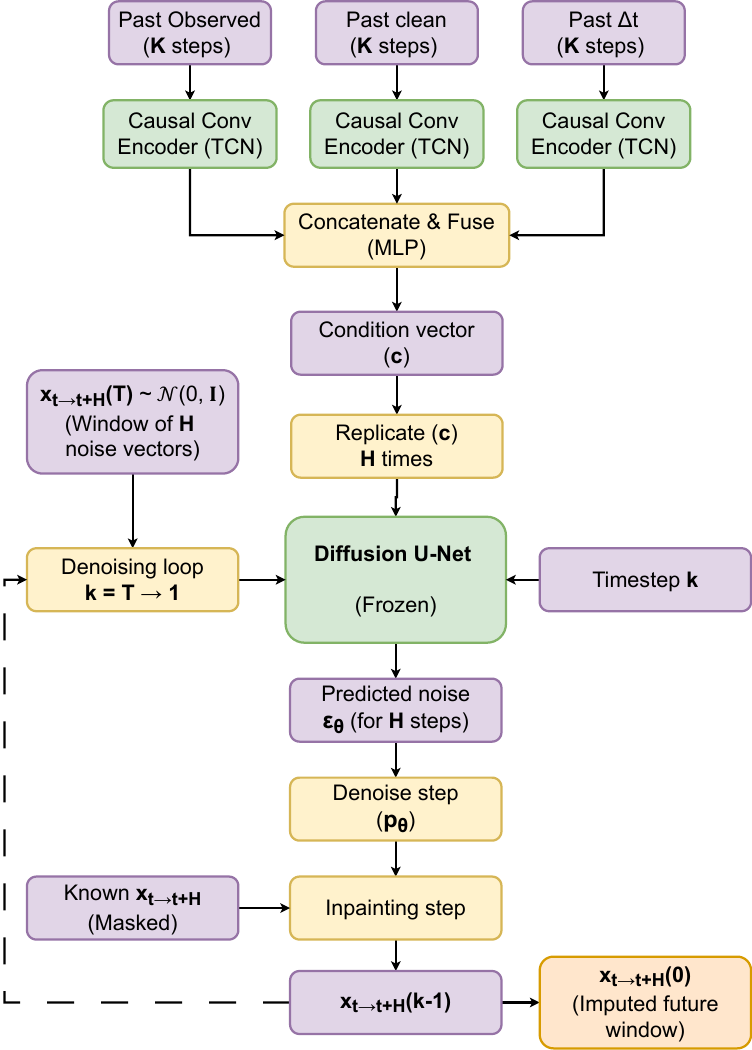}
    \caption{Blockwise imputation with \textbf{STDiff-W}. Given a gap of $H$ missing steps, the model samples an initial noisy window and iteratively denoises it while conditioning on the recent context and clamping any known values inside the block. For longer gaps, the context window is slid forward and successive blocks are imputed and stitched together, combining long-range consistency with locally coherent inpainting.}
    \label{fig:stdiffw_infer}
\end{figure}

\subsubsection{Architecture}
The STDiff-W architecture includes a multi-channel context encoder. It ingests three parallel sequences of length $K$: the last $K$ clean observed states (when simulating training gaps), the corresponding binary mask or observed values (to indicate which of those $K$ steps were actually observed vs. masked), and the time intervals $\Delta t$ between those steps (to inform the model of any irregular sampling). Each sequence is processed by a causal dilated CNN (TCN) with increasing dilation, allowing the context to span long histories. The outputs of these parallel encoders are concatenated and fused (e.g., via a small MLP) to form the final context vector $c$. The context $c$ is then replicated $H$ times (once per future step) and provided to the diffusion U-Net as conditioning at each time step of the $H$-step block. The diffusion U-Net (with frozen weights shared from STDiff) predicts a stack of $H$ noise vectors $\epsilon_\theta$ (one per future step). Figure~\ref{fig:stdiffw_arch} illustrates the training architecture of STDiff-W, where the diffusion U-Net denoises a window of $H$ future steps conditioned on a causal TCN context. 
During inference, the same process operates in reverse - starting from Gaussian noise and iteratively denoising a block of $H$ steps while conditioning on the context vector $c$. 
This inference procedure is shown in Figure~\ref{fig:stdiffw_infer}.

\subsubsection{Inference}
For gaps longer than $H$, we slide the context window and impute successive blocks (optionally with overlap to smooth boundaries). When $K=0$ and $H=1$, STDiff-W reduces to a one-step model akin to STDiff. With large $K$, the design resembles conditional inpainting used by diffusion imputers such as CSDI \citep{tashiro2021csdi}. In practice, we use a fixed context length of $K=128$ steps and a block size of $H=3$ steps. 
This provides a sufficient history for the encoder to capture regime information, while keeping each diffusion denoising pass lightweight and stable. If a gap exceeds $H$, STDiff-W will fill it in multiple stages: the model imputes the first $H$ steps of the gap using the context, then it shifts the window forward (including the newly imputed part as context if treating it as observed for subsequent steps) and imputes the next block, and so on until the gap is filled. This multi-stage strategy allows STDiff-W to handle gaps longer than its block length.

\subsubsection{Irregular sampling and partial observability}
In our implementation, we assume uniform sampling with $\Delta t = 1$, but the framework can incorporate variable elapsed-time features when available. 
Control inputs $u_t$ and exogenous inputs $w_t$ can be partially observed, and we apply masking accordingly. 
This follows established best practices showing that explicit masking and time-gap encoding improve robustness to missingness \citep{che2018grud}.

\subsection{Training objectives and implementation details}
We train the diffusion models (STDiff and STDiff-W) with the standard DDPM $\epsilon$-prediction objective, averaging the mean squared error between the predicted noise and true noise over all steps (for STDiff-W, this average is taken over all $H$ steps in the block). We use an $L^2$ loss:
\[ \mathcal{L}_{\text{diff}} = \mathbb{E}_{x_t,\epsilon,k} \left\|\epsilon - \epsilon_\theta(x_t^{(k)}, k, c_t)\right\|^2, \]
where $x_t^{(k)}$ is the noisy state at step $k$ and $c_t$ is the context.

In principle, repeatedly sampling from the learned reverse-diffusion process yields an ensemble of plausible imputations for each gap, which can be used to construct empirical uncertainty bands (e.g., quantiles or predictive intervals). In this work, however, we focus on point imputations and report MAE/RMSE on single-sample reconstructions for all methods to ensure comparability with prior literature. A systematic treatment of uncertainty quantification and its use in WWTP decision support is left to future work.

Our training protocol for windowed baselines (like SAITS, BRITS when run in window mode) follows standard practice: we set the window length (e.g., $100$ steps for 2-minute resolution data) and slide it over each sequence, imputing within windows and merging overlapping predictions at test time by averaging \citep{pypotsdocs}. Baselines are configured using the PyPOTS library where applicable, to ensure fairness (consistent data handling, mask generation, etc.).

\subsection{Relation to prior diffusion imputers}
Prior diffusion imputers such as CSDI \citep{tashiro2021csdi} treat missing values as undifferentiated masks, ignoring temporal causality and control dynamics. \textsc{STDiff} reframes imputation as probabilistic state-transition simulation, explicitly learning conditional dynamics 
$p(x_t \mid x_{t-1}, u_t, w_t)$. Its windowed variant, \textsc{STDiff-W}, extends this to blockwise denoising with causal context, enabling coherent long-horizon recovery during structured outages. This shift from mask-based filling to control-aware state transitions anchors imputation in the actual process dynamics of industrial systems.

\section{Experiments}

We evaluate the proposed methods on two real-world WWTP datasets under high missingness, using both synthetic and natural gap scenarios. We describe the datasets and masking strategies, baseline methods for comparison, and key evaluation metrics. All experiments are implemented in Python and run on an NVIDIA RTX PRO 6000 GPU.

\subsection{Datasets}
\subsubsection{Agtrup WWTP (primary, quantitative)}
We use a full-scale dataset from Agtrup WWTP (BlueKolding, Denmark). The public release provides a high-frequency SCADA export with 2-minute sampling across two years (Jan 2019-Dec 2020) \citep{mohammadi2024agtrup,mendeley2024agtrup}. In our pipeline, we standardize variables, unify sampling to 2~minutes (upsampling where raw logging is 1-5 minutes, as specified by the authors), and remove obvious telemetry glitches before applying synthetic masking on top of the cleaned series for controlled evaluation as described in Section~4.3.

\subsubsection{Aved\o re WWTP (qualitative stress test with natural outages)}
We additionally analyze a second full-scale dataset from Aved\o re WWTP (Copenhagen, Denmark). The data spans June 2022 - June 2024 with \emph{24 signals} (chemical sensors NH$_4$, NO$_3$, PO$_4$, N$_2$O, DO; flows; temperature; valve positions; blower airflow; phase codes) \citep{hansen2024n2o}. Crucially, this set contains real sensor outages (from fouling, maintenance, etc.) for which ground truth is unavailable. We use this dataset primarily to qualitatively assess how each method fills long, irregular gaps under real conditions.

\subsubsection{Why these datasets are a strong testbed}
(1) \textit{Full-scale operations with rich covariates.} 
Both sites provide dense SCADA logging, including actuator commands and phase indicators, enabling conditional models to exploit causal cues \citep{mohammadi2024agtrup,hansen2024n2o}. 

(2) \textit{Complex dynamics and regime shifts.} 
Industrial WWTP data exhibit nonlinear and regime-dependent behavior, mixing diurnal control oscillations with seasonal and shock-driven variability. Such multi-scale dynamics challenge both local pattern learners and long-range simulators to preserve physical realism \citep{martin2014influent,epa1999sbr,zhi2024deeplearningforwaterquality}.

(3) \textit{Control-driven regime switches.} 
Phase changes and setpoint adjustments induce abrupt regime shifts in certain variables (e.g., aeration phases causing oscillatory $NH_4$ patterns). These aspects make the datasets ideal for testing imputers on both long-range consistency and short-term detail preservation \citep{epa1999sbr,zhi2024deeplearningforwaterquality}.

\subsection{Experimental setup}
We simulate high-missingness scenarios representative of plant operations. The timeline is split into \textbf{70\% train}, \textbf{15\% validation}, and \textbf{15\% test} (the test set is the last 15\% of the series). We corrupt the data synthetically (block + pointwise gaps). PyPOTS baselines are trained on the \emph{corrupted} train windows and validated on the \emph{corrupted} validation windows, while being evaluated on the corrupted test windows against the clean test targets. \textbf{STDiff} (one-step) is trained on step-ahead pairs constructed from the synthetically corrupted training slice and is used to inpaint only the next-state entries at test time.

\textbf{STDiff-W} uses clean past features with their aligned corrupted counterparts to form its conditioning. During training, \textbf{STDiff-W} receives both the clean and corrupted versions of each past window. The clean segment provides ground-truth supervision for the denoising loss, while the corrupted version serves as the model’s input context. This setup mirrors self-supervised denoising autoencoder training: the model learns to reconstruct clean sequences from synthetically masked counterparts. 
At test time, only the corrupted inputs are provided. Hyperparameters are fixed \emph{a priori} and model selection relies on \emph{early stopping} on the validation split; no large hyperparameter sweep is performed.

\begin{table}[!t]
\centering
\caption{Per-feature realized missingness (\%) at each nominal masking level (Agtrup).}
\label{tab:miss_all}
\tblsetup 
\begin{tabular}{@{}l
  S[table-format=2.2]
  S[table-format=2.2]
  S[table-format=2.2]
  S[table-format=2.2]@{}}
\toprule
\textbf{Feature} & \textbf{20\%} & \textbf{30\%} & \textbf{40\%} & \textbf{50\%} \\
\midrule
T1\_NH4               & 11.54 & 17.25 & 22.66 & 28.09 \\
T1\_PO4               & 11.61 & 17.20 & 22.67 & 28.48 \\
IN\_METAL\_Q          &  5.47 &  8.36 & 11.12 & 13.75 \\
T1\_O2                &  5.59 &  8.28 & 11.07 & 13.74 \\
METAL\_Q              &  5.55 &  8.22 & 10.97 & 13.76 \\
TEMPERATURE           &  5.55 &  8.38 & 11.12 & 13.65 \\
IN\_Q                 &  5.56 &  8.28 & 11.11 & 13.70 \\
MAX\_CF               &  5.57 &  8.23 & 11.03 & 13.74 \\
PROCESSPHASE\_INLET   &  5.61 &  8.31 & 11.02 & 13.68 \\
PROCESSPHASE\_OUTLET  &  5.60 &  8.28 & 10.96 & 13.56 \\
\bottomrule
\end{tabular}
\end{table}

\begin{table}[!tb]
\centering
\caption{Final imputation metrics on Agtrup (RAW and Z). Corr = nominal corruption rate.}
\label{tab:imputation_full_unified}
\tblsetup                               
\setlength{\tabcolsep}{4pt}             
\begin{tabular}{@{}%
  L{0.17\textwidth}  
  c                  
  S[table-format=1.4] S[table-format=1.4]  
  S[table-format=1.4] S[table-format=1.4]  
  S[table-format=1.4] S[table-format=1.4]  
  S[table-format=1.4] S[table-format=1.4]  
@{}}
\toprule
\multirow{2}{*}{\textbf{Model}} & \multirow{2}{*}{\textbf{Corr}} &
\multicolumn{2}{c}{\textbf{NH$_4$ (raw)}} &
\multicolumn{2}{c}{\textbf{PO$_4$ (raw)}} &
\multicolumn{2}{c}{\textbf{NH$_4$ (z)}} &
\multicolumn{2}{c}{\textbf{PO$_4$ (z)}} \\
 &  & {\textbf{MAE}} & {\textbf{RMSE}} & {\textbf{MAE}} & {\textbf{RMSE}} &
      {\textbf{MAE}} & {\textbf{RMSE}} & {\textbf{MAE}} & {\textbf{RMSE}} \\
\midrule
SAITS         & 20\% & 0.2340 & 0.5452 & 0.1325 & 0.2575 & 0.1441 & 0.3358 & 0.1803 & 0.3503 \\
iTransformer  & 20\% & 0.3498 & 0.6301 & 0.1847 & 0.3096 & 0.2155 & 0.3881 & 0.2513 & 0.4213 \\
BRITS         & 20\% & 0.1571 & 0.3220 & 0.1196 & 0.2205 & 0.0967 & 0.1983 & 0.1627 & 0.3001 \\
CSDI          & 20\% & 0.2134 & 0.5183 & 0.1558 & 0.3411 & 0.1314 & 0.3192 & 0.2120 & 0.4641 \\
STDiff        & 20\% & 0.2952 & 0.6637 & 0.1730 & 0.3416 & 0.1818 & 0.4087 & 0.2354 & 0.4648 \\
STDiff-W      & 20\% & 0.1581 & 0.2545 & 0.1044 & 0.2057 & 0.0973 & 0.1568 & 0.1421 & 0.2800 \\
Latent ODE    & 20\% & 0.6321 & 0.8584 & 0.4070 & 0.5587 &  0.3910 & 0.5295 & 0.5575 & 0.7624 \\
TimeMixer++   & 20\% & 0.8692 & 1.0375 & 0.4894 & 0.6481 & 0.5353 & 0.6390 & 0.6659 & 0.8819 \\
\midrule
SAITS         & 30\% & 0.2108 & 0.3776 & 0.1284 & 0.2676 & 0.1298 & 0.2325 & 0.1747 & 0.3642 \\
iTransformer  & 30\% & 0.3273 & 0.5181 & 0.1913 & 0.3126 & 0.2016 & 0.3191 & 0.2603 & 0.4254 \\
BRITS         & 30\% & 0.1571 & 0.2920 & 0.1035 & 0.1886 & 0.0967 & 0.1799 & 0.1409 & 0.2566 \\
CSDI          & 30\% & 0.2178 & 0.4287 & 0.1489 & 0.3073 & 0.1341 & 0.2640 & 0.2027 & 0.4182 \\
STDiff        & 30\% & 0.3202 & 0.7011 & 0.1927 & 0.3699 & 0.1972 & 0.4318 & 0.2622 & 0.5034 \\
STDiff-W      & 30\% & 0.1556 & 0.2604 & 0.0973 & 0.1889 & 0.0958 & 0.1604 & 0.1324 & 0.2571 \\
Latent ODE    & 30\% & 0.6113 & 0.8029 & 0.3984 & 0.5444 &  0.3758 & 0.4927 & 0.5359 & 0.7373 \\
TimeMixer++   & 30\% & 0.8936 & 1.0614 & 0.4791 & 0.6155 & 0.5503 & 0.6537 & 0.6520 & 0.8375 \\
\midrule
SAITS         & 40\% & 0.2063 & 0.3903 & 0.1581 & 0.2913 & 0.1270 & 0.2404 & 0.2152 & 0.3964 \\
iTransformer  & 40\% & 0.3355 & 0.5315 & 0.2103 & 0.3325 & 0.2066 & 0.3273 & 0.2862 & 0.4525 \\
BRITS         & 40\% & 0.1603 & 0.2910 & 0.1256 & 0.2642 & 0.0988 & 0.1792 & 0.1709 & 0.3595 \\
CSDI          & 40\% & 0.2202 & 0.4727 & 0.1640 & 0.3353 & 0.1356 & 0.2911 & 0.2232 & 0.4562 \\
STDiff        & 40\% & 0.3324 & 0.6927 & 0.2137 & 0.4095 & 0.2047 & 0.4266 & 0.2908 & 0.5572 \\
STDiff-W      & 40\% & 0.1479 & 0.2538 & 0.0959 & 0.1893 & 0.0911 & 0.1563 & 0.1305 & 0.2576 \\
Latent ODE    & 40\% & 0.6114 & 0.7991 & 0.4067 & 0.5525 & 0.3759 & 0.4917 & 0.5516 & 0.7507 \\
TimeMixer++   & 40\% & 0.9781 & 1.1673 & 0.5278 & 0.6773 & 0.6024 & 0.7189 & 0.7181 & 0.9217 \\
\midrule
SAITS         & 50\% & 0.2033 & 0.3549 & 0.1658 & 0.3471 & 0.1252 & 0.2185 & 0.2256 & 0.4723 \\
iTransformer  & 50\% & 0.3457 & 0.5390 & 0.2210 & 0.3726 & 0.2129 & 0.3319 & 0.3007 & 0.5070 \\
BRITS         & 50\% & 0.1559 & 0.2773 & 0.1307 & 0.2584 & 0.0960 & 0.1708 & 0.1778 & 0.3516 \\
CSDI          & 50\% & 0.2298 & 0.4779 & 0.1738 & 0.3609 & 0.1416 & 0.2943 & 0.2365 & 0.4911 \\
STDiff        & 50\% & 0.3678 & 0.8064 & 0.2410 & 0.4634 & 0.2265 & 0.4967 & 0.3280 & 0.6306 \\
STDiff-W      & 50\% & 0.1437 & 0.2369 & 0.0938 & 0.1905 & 0.0885 & 0.1459 & 0.1276 & 0.2593 \\
Latent ODE    & 50\% & 0.5991 & 0.7939 & 0.4027 & 0.5672 & 0.3695 & 0.4899 & 0.5476 & 0.7713 \\
TimeMixer++   & 50\% & 0.8529 & 1.0797 & 0.4786 & 0.6500 & 0.5253 & 0.6650 & 0.6512 & 0.8846 \\
\bottomrule
\end{tabular}
\end{table}

\subsection{Missingness simulation on Agtrup}
We evaluate at nominal missingness levels $\{20\%, 30\%, 40\%, 50\%\}$. Each mask combines contiguous block outages and scattered pointwise drops, reflecting both long fouling outages and random telemetry glitches. Specifically, we randomly remove segments of varying lengths (drawn from a distribution to mimic real outages) to constitute the majority of missing data, and additionally remove individual points at random to reach the desired overall percentage. This design follows recommendations to move beyond MCAR and stress models under contiguous removals \citep{khayati2020mindthegap}. The same masks are applied across all methods for fair comparison. Realized missing fractions per variable range between the specified rates (e.g., 20-50\%) due to different gap placements (see Table ~\ref{tab:miss_all}  for exact values).

We deliberately cap the synthetic missingness at 50\%. Imputing under heavier corruption (\(>50\%\)) is known to be extremely challenging: models often converge to trivial or nearly flat solutions over long continuous gaps, and attention-based methods can underperform because insufficient temporal context remains inside each gap \citep{chari2025simulatedannealing,niako2024bloodpressure}. Moreover, most empirical studies on time-series imputation and forecasting do not evaluate beyond roughly 30-50\% missingness for similar reasons \citep{khayati2020mindthegap,niako2024bloodpressure}. To still examine more extreme and irregular scenarios, we complement these synthetic experiments with qualitative analyses on the Aved\o re dataset, which contains long natural sensor outages that go beyond the controlled blackout patterns used in our synthetic masks.

\subsubsection{Downstream forecasting evaluation}
To assess utility beyond point error, we train a simple one-step forecaster \emph{separately for each method} using that method’s \emph{imputed training set}, and then evaluate forecasting MAE/RMSE on the test horizon using the same method’s imputed test inputs. This task-oriented evaluation isolates how well each imputer supports downstream prediction, not only reconstruction accuracy under masking. Full numeric results (RAW units) are reported in Table ~\ref{tab:forecast_full_unified}.

\subsubsection{Causal-style ablations on conditioning signals}
We assess whether \textsc{STDiff} actually \emph{uses} control and exogenous inputs by re-evaluating a \emph{trained} model on \emph{test} under input perturbations applied \emph{only at inference}: 
\emph{drop control} (zero the control channels), 
\emph{drop exogenous} (zero the exogenous channels), 
\emph{drop both}, 
\emph{shuffle control} (batch permutation of control channels), 
\emph{shuffle exogenous}, 
\emph{noise control} (replace with $\mathcal{N}(0,1)$), and 
\emph{noise exogenous}. 
For each nominal corruption level (20-50\%), we recompute imputations and report NH$_4$ and PO$_4$ MAE/RMSE (raw units) at the originally missing next-state positions. 
Ablations are conducted for \textsc{STDiff} (one-step) only. 
Results are summarized in Table ~\ref{tab:ablations_raw}; the largest degradations typically occur when exogenous inputs are removed.

\subsection{Compared methods and fairness controls}
We compare against representative baselines spanning deterministic/probabilistic imputers and diverse sequence inductive biases: \textbf{SAITS} (Transformer imputer with diagonal self-attention) \citep{du2023saits}, \textbf{BRITS} (bidirectional RNN with learnable missing values) \citep{cao2018brits}, \textbf{iTransformer} (inverted-dimension Transformer used widely for forecasting and adapted in PyPOTS) \citep{liu2023itransformer}, \textbf{TimeMixer} (multiscale MLP mixing) \citep{wang2024timemixer}, \textbf{Latent ODE} (continuous-time latent dynamics) \citep{rubanova2019latentode}, and \textbf{CSDI} (conditional diffusion imputer) \citep{tashiro2021csdi}.
All baselines are trained on synthetically corrupted TRAIN windows, validated on corrupted VAL windows, and evaluated on corrupted TEST windows against the clean TEST targets.

\subsubsection{Tooling and windowing.}
PyPOTS models use a sliding-window API; we set $n_{\text{steps}}{=}100$. 
On TEST, we use \emph{non-overlapping} windows (stride $= n_{\text{steps}}$) and place each window’s reconstruction at its location without boundary averaging.

\subsection{\textsc{STDiff}/\textsc{STDiff-W} settings}
\textsc{STDiff} conditions a DDPM denoiser on $[x_{t-1};u_t;w_t]$ and is used as a \emph{one-step} imputer: it predicts $x_{t}$ from $(x_{t-1},u_t,w_t)$ and inpaints only the next-state missing entries (no multi-step rollouts through gaps).

\textsc{STDiff-W} augments \textsc{STDiff} with a causal window encoder over the last $K$ steps and performs block-wise denoising of length $H$ at inference. In our implementation, \textsc{STDiff-W} is trained with the standard \(\epsilon\)-prediction objective applied jointly over a block of \(H\) future steps. During inference, the context vector is replicated across the \(H\) steps, and any known values within the block are clamped. We use \(K{=}128\) and \(H{=}3\) in all experiments.

\subsection{Metrics and reporting}
We report MAE and RMSE on masked entries (imputed positions only) and the downstream one-step forecasting MAE and RMSE obtained by training the forecaster on clean TRAIN data and evaluating on TEST with each method’s imputed inputs.

\section{Results}
\label{sec:results}

\subsection{Quantitative imputation on synthetic gaps (Agtrup)}
We evaluate block-heavy masking (plus scattered points) at nominal rates $\{20,30,40,50\}\%$ applied identically across methods. We report MAE/RMSE in raw units for NH$_4$ and PO$_4$; z-score trends are consistent (Table~\ref{tab:imputation_full_unified}).

\textbf{STDiff-W} attains the best overall reconstruction across all missingness levels and both targets. 
At \textbf{50\%} masking it yields the lowest errors (NH$_4$: 0.1437/0.2369; PO$_4$: 0.0938/0.1905), outperforming \textbf{BRITS}, \textbf{SAITS}, \textbf{CSDI}, and \textbf{iTransformer}. 
At lighter masks (20--30\%), BRITS is very competitive on NH$_4$ \emph{MAE} but \textbf{STDiff-W} keeps a clear edge on \emph{RMSE} and is consistently best on PO$_4$.

\subsubsection{Method-wise trends}
\begin{itemize}
    \item \textbf{STDiff-W (windowed diffusion)}: lowest errors overall; sublinear degradation with missingness. Context helps curb drift on long gaps.
    \item \textbf{STDiff (one-step diffusion)}: clearly below STDiff-W and degrades with missingness, reflecting error accumulation without windowed context.
    \item \textbf{BRITS}: strong at 20-30\% (esp.\ NH$_4$ MAE) but loses ground as missingness grows.
    \item \textbf{SAITS} and \textbf{iTransformer}: competitive at lighter masks; plateau under blackout-dominated regimes.
    \item \textbf{TimeMixer++} and \textbf{Latent ODE}: substantially higher errors under these masking patterns.
\end{itemize}

\textbf{STDiff-W} ranks first on reconstruction; BRITS/SAITS form a middle tier (rate/variable dependent); \textbf{STDiff} (one-step), iTransformer, and others trail at higher missing rates.

\subsection{Downstream one-step forecasting (task-oriented utility)}
We evaluate how well each imputation method supports prediction using a simple one-step forecaster (Table ~\ref{tab:forecast_full_unified}). The forecaster is trained once on clean \textsc{Train} data and, for each imputer and corruption level, evaluated on that imputer’s imputed \textsc{Test} series. This design avoids train-test distribution mismatch, isolates whether an imputer preserves predictive structure (e.g., lags, peaks, lead-lag relations), and makes differences attributable to the imputed inputs rather than to re-training the forecaster. Clear trends emerge across missingness levels for NH$_4$ and PO$_4$, as detailed below.

\begin{table}[!t]
\centering
\caption{Downstream one-step forecasting MAE/RMSE (RAW units) using each method’s imputed training set.}
\label{tab:forecast_full_unified}
\begingroup
\setlength{\tabcolsep}{4pt}      
\renewcommand{\arraystretch}{1.15}
\begin{tabular}{@{}%
  L{0.15\textwidth}  
  c                   
  S[table-format=1.4] S[table-format=1.4]  
  S[table-format=1.4] S[table-format=1.4]  
@{}}
\toprule
\multirow{2}{*}{\textbf{Model}} & \multirow{2}{*}{\textbf{Corr}} &
\multicolumn{2}{c}{\textbf{NH$_4$}} &
\multicolumn{2}{c}{\textbf{PO$_4$}} \\
 &  & \textbf{MAE} & \textbf{RMSE} & \textbf{MAE} & \textbf{RMSE} \\
\midrule
SAITS         & 20\% & 0.1359 & 0.2709 & 0.0937 & 0.1980 \\
iTransformer  & 20\% & 0.1495 & 0.2917 & 0.0994 & 0.2066 \\
BRITS         & 20\% & 0.1278 & 0.2272 & 0.0919 & 0.1926 \\
CSDI          & 20\% & 0.1348 & 0.2649 & 0.0965 & 0.2106 \\
STDiff        & 20\% & 0.1471 & 0.3029 & 0.1014 & 0.2182 \\
STDiff-W      & 20\% & 0.1309 & 0.2220 & 0.0940 & 0.1981 \\
Latent ODE    & 20\% & 0.1807 & 0.3535 & 0.1231 & 0.2580 \\
TimeMixer++   & 20\% & 0.2063 & 0.4034 & 0.1339 & 0.2822 \\
\midrule
SAITS         & 30\% & 0.1389 & 0.2467 & 0.0965 & 0.2057 \\
iTransformer  & 30\% & 0.1590 & 0.2884 & 0.1066 & 0.2161 \\
BRITS         & 30\% & 0.1300 & 0.2253 & 0.0918 & 0.1892 \\
CSDI          & 30\% & 0.1411 & 0.2600 & 0.1002 & 0.2136 \\
STDiff        & 30\% & 0.1627 & 0.3528 & 0.1119 & 0.2393 \\
STDiff-W      & 30\% & 0.1347 & 0.2290 & 0.0957 & 0.1995 \\
Latent ODE    & 30\% & 0.2064 & 0.3864 & 0.1405 & 0.2833 \\
TimeMixer++   & 30\% & 0.2487 & 0.4747 & 0.1521 & 0.3054 \\
\midrule
SAITS         & 40\% & 0.1432 & 0.2632 & 0.1060 & 0.2179 \\
iTransformer  & 40\% & 0.1718 & 0.3135 & 0.1175 & 0.2320 \\
BRITS         & 40\% & 0.1339 & 0.2330 & 0.0981 & 0.2095 \\
CSDI          & 40\% & 0.1479 & 0.2910 & 0.1080 & 0.2334 \\
STDiff        & 40\% & 0.1775 & 0.3830 & 0.1250 & 0.2677 \\
STDiff-W      & 40\% & 0.1375 & 0.2342 & 0.0989 & 0.2044 \\
Latent ODE    & 40\% & 0.2323 & 0.4245 & 0.1615 & 0.3144 \\
TimeMixer++   & 40\% & 0.3134 & 0.5862 & 0.1848 & 0.3599 \\
\midrule
SAITS         & 50\% & 0.1490 & 0.2642 & 0.1127 & 0.2458 \\
iTransformer  & 50\% & 0.1871 & 0.3396 & 0.1275 & 0.2569 \\
BRITS         & 50\% & 0.1350 & 0.2350 & 0.1017 & 0.2114 \\
CSDI          & 50\% & 0.1576 & 0.3134 & 0.1158 & 0.2523 \\
STDiff        & 50\% & 0.2024 & 0.4695 & 0.1410 & 0.3050 \\
STDiff-W      & 50\% & 0.1408 & 0.2361 & 0.1015 & 0.2080 \\
Latent ODE    & 50\% & 0.2573 & 0.4628 & 0.1787 & 0.3448 \\
TimeMixer++   & 50\% & 0.3331 & 0.6107 & 0.1950 & 0.3776 \\
\bottomrule
\end{tabular}
\endgroup
\end{table}

\begin{table}[!t]
\centering
\caption{STDiff causal ablations (raw units) on Agtrup across nominal missingness rates. Corr = nominal corruption rate.}
\label{tab:ablations_raw}
\tblsetup                                 
\setlength{\tabcolsep}{4pt}               
\renewcommand{\arraystretch}{1.15}        

\begin{tabular}{@{}%
  c                       
  L{0.32\linewidth}       
  S[table-format=1.4]     
  S[table-format=1.4]     
  S[table-format=1.4]     
  S[table-format=1.4]     
@{}}
\toprule
\multicolumn{2}{@{}l}{} &
\multicolumn{2}{c}{\textbf{NH$_4$}} &
\multicolumn{2}{c}{\textbf{PO$_4$}} \\
\cmidrule(lr){3-4}\cmidrule(lr){5-6}
\textbf{Corr} & \textbf{Ablation} & \textbf{MAE} & \textbf{RMSE} & \textbf{MAE} & \textbf{RMSE} \\
\midrule
20\% & none              & 0.3161 & 0.6953 & 0.2006 & 0.4079 \\
20\% & drop control      & 0.3279 & 0.7110 & 0.2085 & 0.4193 \\
20\% & drop exogenous    & 0.4134 & 0.9245 & 0.2147 & 0.4349 \\
20\% & drop both         & 0.4198 & 0.9193 & 0.2202 & 0.4374 \\
20\% & shuffle control   & 0.3208 & 0.6972 & 0.2076 & 0.4178 \\
20\% & shuffle exogenous & 0.3443 & 0.7513 & 0.2085 & 0.4190 \\
20\% & noise control     & 0.3358 & 0.7295 & 0.2071 & 0.4134 \\
20\% & noise exogenous   & 0.4302 & 0.9143 & 0.2124 & 0.4247 \\
\midrule
30\% & none              & 0.3347 & 0.7006 & 0.2147 & 0.4098 \\
30\% & drop control      & 0.3418 & 0.7033 & 0.2211 & 0.4230 \\
30\% & drop exogenous    & 0.4123 & 0.8814 & 0.2217 & 0.4294 \\
30\% & drop both         & 0.4368 & 0.9132 & 0.2317 & 0.4429 \\
30\% & shuffle control   & 0.3379 & 0.6885 & 0.2248 & 0.4330 \\
30\% & shuffle exogenous & 0.3772 & 0.8038 & 0.2270 & 0.4348 \\
30\% & noise control     & 0.3483 & 0.7084 & 0.2231 & 0.4277 \\
30\% & noise exogenous   & 0.4423 & 0.9071 & 0.2246 & 0.4344 \\
\midrule
40\% & none              & 0.3475 & 0.7122 & 0.2398 & 0.4545 \\
40\% & drop control      & 0.3651 & 0.7415 & 0.2477 & 0.4666 \\
40\% & drop exogenous    & 0.4410 & 0.8897 & 0.2554 & 0.4818 \\
40\% & drop both         & 0.4560 & 0.8899 & 0.2749 & 0.5132 \\
40\% & shuffle control   & 0.3511 & 0.6980 & 0.2429 & 0.4613 \\
40\% & shuffle exogenous & 0.3915 & 0.7748 & 0.2539 & 0.4744 \\
40\% & noise control     & 0.3650 & 0.7142 & 0.2480 & 0.4680 \\
40\% & noise exogenous   & 0.4914 & 0.9564 & 0.2568 & 0.4848 \\
\midrule
50\% & none              & 0.3851 & 0.7973 & 0.2646 & 0.4990 \\
50\% & drop control      & 0.3974 & 0.8142 & 0.2842 & 0.5286 \\
50\% & drop exogenous    & 0.4851 & 0.9572 & 0.2790 & 0.5251 \\
50\% & drop both         & 0.4880 & 0.9572 & 0.3007 & 0.5526 \\
50\% & shuffle control   & 0.3897 & 0.7944 & 0.2661 & 0.5085 \\
50\% & shuffle exogenous & 0.4344 & 0.8584 & 0.2816 & 0.5332 \\
50\% & noise control     & 0.4015 & 0.8080 & 0.2828 & 0.5300 \\
50\% & noise exogenous   & 0.5291 & 1.0176 & 0.2831 & 0.5373 \\
\bottomrule
\end{tabular}
\end{table}

\subsubsection{NH$_4$}
\textbf{BRITS} and \textbf{STDiff-W} consistently achieve the strongest forecasting accuracy, substantially outperforming the other imputers.  
At \textbf{20\%} missingness, BRITS attains \textbf{MAE = 0.1278}, \textbf{RMSE = 0.2272}, while STDiff-W achieves \textbf{MAE = 0.1309}, \textbf{RMSE = 0.2220} (very similar MAE, slightly lower RMSE).  
Even at \textbf{50\%}, the two remain neck-and-neck: BRITS \textbf{0.1350 / 0.2350} vs.\ STDiff-W \textbf{0.1408 / 0.2361}. STDiff-W degrades only slightly and stays within $\approx 0.005$ MAE of BRITS at 50\%. No other method is close on NH$_4$ (e.g., SAITS MAE $\approx 0.149$ at 50\%).

\subsubsection{PO$_4$}
\textbf{STDiff-W} is top or statistically tied for top across all rates.  
At \textbf{50\%} missingness, STDiff-W leads with \textbf{MAE = 0.1015}, \textbf{RMSE = 0.2080}, slightly better than BRITS (\textbf{0.1017 / 0.2114}).  
At \textbf{20\%}, STDiff-W’s \textbf{MAE = 0.0940} vs.\ BRITS \textbf{0.0919} (gap $< 0.002$), effectively a tie at low missingness; STDiff-W pulls slightly ahead as gaps grow. Other baselines (e.g., SAITS at 50\%: \textbf{MAE = 0.1127}) are weaker.

\subsubsection{Other methods}
The one-step diffusion baseline \textbf{STDiff} (without windowed context) underperforms at all rates (e.g., NH$_4$ at 50\%: \textbf{0.2024 / 0.4695}), underscoring the benefit of STDiff-W’s context conditioning. CSDI, iTransformer, and TimeMixer also lag - typically by 10-50\% relative error margins versus STDiff-W.

\subsubsection{Summary}
Imputation quality and forecasting quality are related but not identical. \textbf{STDiff-W} provides state-of-the-art imputations \emph{and} strong predictive performance; \textbf{BRITS} can match or slightly beat it on NH$_4$ at lighter missingness despite higher reconstruction error. This highlights that imputers should be judged by \emph{downstream impact}, not MAE/RMSE alone: preserving dynamics (even with slightly higher point error) can yield better decision-support in industrial systems.

\subsection{Causal-style ablations (controls/exogenous)}
Ablating inputs for a trained \textbf{STDiff} at inference time (\texttt{drop}/\texttt{shuffle}/\texttt{noise}) shows the largest degradations when \emph{exogenous} channels are removed or noised (e.g., NH$_4$ MAE at 50\%: 0.3851 $\to$ 0.4851/0.5291). This indicates meaningful use of $u_t$/$w_t$ (Table ~\ref{tab:ablations_raw}).



\subsection{Qualitative evaluation on natural outages (Aved\o re)}

\begin{figure*}
  \centering
  \includegraphics[width=\textwidth,height=0.88\textheight,keepaspectratio]{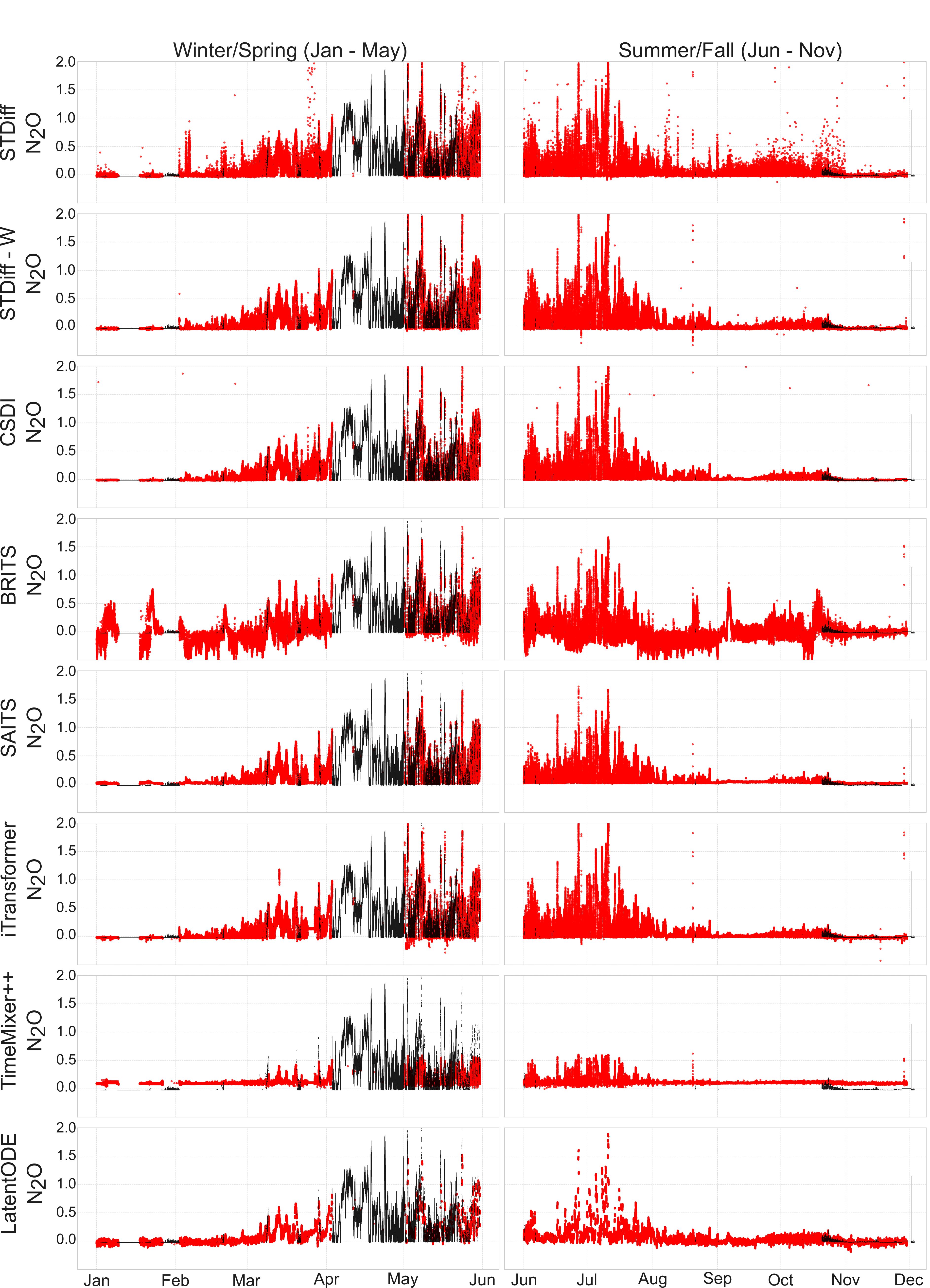}
  \caption{N$_2$O concentration at Aved\o re WWTP during natural sensor outages in winter/spring (left) and summer/fall (right). Each row shows the same missing segments reconstructed by a different model, illustrating how methods differ in preserving oscillations, spikes, and regime changes under long, irregular gaps.}
  \label{fig:n2o_imputation}
\end{figure*}

\begin{figure*}
  \centering
  \includegraphics[width=\textwidth,height=0.90\textheight,keepaspectratio]{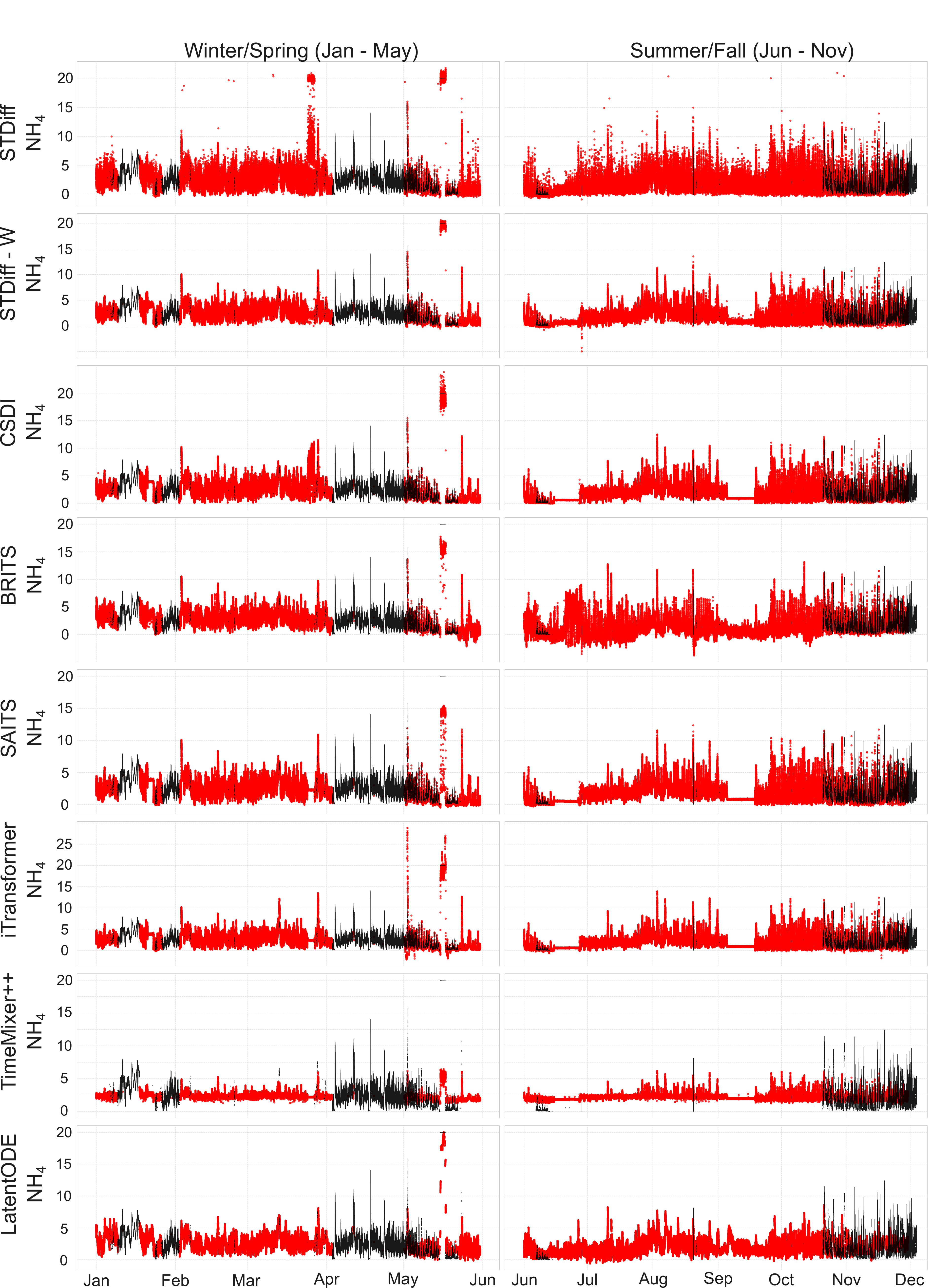}
  \caption{NH$_4$ concentration at Aved\o re WWTP during natural sensor outages in winter/spring (left) and summer/fall (right). The panels compare how different imputers fill multi-day gaps, highlighting differences in amplitude preservation, phase alignment with operational cycles, and the treatment of sharp regime shifts.}
  \label{fig:nh4_imputation}
\end{figure*}

Without ground truth for real outages, we assess the visual fidelity and consistency of different methods’ imputations. We focus on two example signals with clear patterns: nitrous oxide ($N_2O$) and ammonium ($NH_4$) during known aeration cycles (Figure ~\ref{fig:n2o_imputation} \& Figure ~\ref{fig:nh4_imputation}).

\textbf{Regime oscillation fidelity:} STDiff-W convincingly reproduces the amplitude-modulated oscillations that align with the plant’s aeration on/off phases, and it stitches gap boundaries smoothly into the observed context. In contrast, SAITS and iTransformer tend to flatten the signal toward a mid-range level over multi-day gaps, losing the oscillatory extremes.

\textbf{Extreme events and regime changes:} STDiff-W imputes sharp spikes and surges (e.g., sudden $NH_4$ jumps) in a manner timed with known process events (such as anoxic phase onsets). Many baselines under-shoot these spikes, effectively damping variability (a form of variance attenuation). For instance, where STDiff-W inserts a pronounced peak that matches patterns seen before the gap, baselines often produce a smaller bump or a smooth ramp, indicating they are averaging out the extreme.

Based on visual examination of long outage periods, we would rank the methods roughly as follows for realism: \textsc{STDiff-W} is best (capturing both oscillatory behavior and sudden shifts), followed by \textsc{STDiff} and \textsc{CSDI} (which also maintain visible dynamics fairly well). \textsc{SAITS} and \textsc{iTransformer} preserve broad trends but often attenuate amplitudes and occasionally drift in phase over seasonal cycles. \textsc{TimeMixer} performs comparably or slightly worse: it frequently underestimates spike magnitude and exhibits lead-lag (phase) shifts relative to the observed patterns - visible for NH$_4$ during summer/fall and for N$_2$O peaks. \textsc{BRITS} tends to flatten extended gaps, with occasional boundary discontinuities at gap edges, while \textsc{LatentODE} shows strong smoothing with temporal lag, losing oscillatory structure and shock responses across seasonal transitions. These qualitative rankings mirror the quantitative robustness and downstream results: models that numerically handled block missingness better also yielded visually more plausible reconstructions.

For NH$_4$ specifically, we observed characteristic seasonal patterns: a pronounced spike during late May-June and extended missing segments in June-July and again in September-October. During these summer and early-fall outages, \textsc{CSDI} occasionally produced near-flat imputations, whereas \textsc{STDiff} and \textsc{STDiff-W} maintained realistic within-gap variability. This suggests that, under long contiguous gaps, window-only diffusion conditioning can suffer from context starvation, while transition-based models with per-step control inputs better preserve temporal dynamics.

\section{Discussion}
\label{sec:discussion}

\paragraph{\textbf{Validate on realistic missingness, not only MCAR\\}}
Our results reinforce that evaluation on \emph{structured outages} matters. Methods that look strong under short, pointwise MCAR masks can degrade under contiguous blackouts that mimic maintenance, fouling, or telemetry loss. In our study, windowed baselines (e.g., SAITS) and BRITS were competitive on synthetic gaps yet tended to plateau or over-smooth under long, irregular Aved\o re outages - yielding visually plausible but dynamic-poor fills. This echoes benchmarking advice to include \emph{block masks} and \emph{natural gaps} when assessing imputers \citep{khayati2020mindthegap}. Practically, we recommend reserving a slice of raw operational data as a qualitative testbed alongside synthetic masks.

\paragraph{\textbf{Long-range continuity + short-range detail\\}}
Across regimes, \textsc{STDiff-W} performed best; its blockwise diffusion conditioned on a causal context window preserves local detail while remaining globally coherent. Importantly, \textsc{STDiff} (sequential) does not have an “unlimited receptive field’’ in the attention/RNN sense: each step conditions on $[x_{t-1};u_t]$. What it \emph{does} have is an \emph{unbounded rollout horizon} - it can simulate as long as needed, with its generated $x_{t-1}$ carrying information forward. In practice, this avoids window starvation but can accumulate error over very long gaps. \textsc{STDiff-W} counteracts that by jointly denoising an $H$-step block with a $K$-step context, reducing drift and sharpening short-term structure, consistent with DDPM conditioning behavior \citep{ho2020ddpm}. The general lesson: mechanisms that combine global continuity with local fidelity handle industrial blackouts better than strictly windowed or strictly one-step models. This effect is visible in the Avedøre NH$_4$ series, where \textsc{CSDI} produced flat segments during the June-July and September-October outages, while \textsc{STDiff-W} preserved oscillations within those same gaps. The contrast supports the view that transition-based conditioning alleviates context starvation and maintains finer temporal structure under sustained missingness.

\paragraph{\textbf{Exogenous and causal inputs are not optional\\}}
Conditioning on controls and exogenous signals (valves, phase, flow/temperature) materially improves realism. Our ablations show the largest degradations when exogenous channels were removed or corrupted 
(e.g., NH$_4$ MAE increased by 26-37\% at 50\% masks, depending on the ablation type), 
with control inputs also contributing but less strongly. Qualitatively, \textsc{STDiff-W} aligned spikes and regime changes with phase/actuation in Aved\o re; baselines lacking these cues often flattened or mis-timed events. We recommend practitioners use domain-salient indicators (even simple binaries) to guide imputers away from physically implausible fills.

\paragraph{\textbf{Look beyond MAE/RMSE: visual and task-oriented checks\\}} 
Point error alone can reward variance shrinkage and mean-like fills. In our downstream test, STDiff-W paired top imputation accuracy with top (or tied-top) forecasting, while some over-smoothed methods underperformed downstream despite low MAE - aligning with task-oriented evaluation that prioritizes downstream utility \citep{wang2024downstream}.

Even when an imputer attains low pointwise MAE, its outputs may attenuate lagged correlations or dampen peaks, weakening downstream predictors. As corruption increases, a larger share of the forecaster’s inputs are smoothed imputations, leading to covariate shift and signal shrinkage - effects that our task-oriented evaluation exposes directly.

For deployments, we advocate a triad: (i) error on masked points, (ii) downstream performance on relevant tasks, and (iii) targeted visual audits with domain experts.

\paragraph{\textbf{Non-stationarity and seasonality\\}}
WWTP signals shift across seasons and operating regimes. Our year-spanning results suggests \textsc{STDiff-W} copes well (aided by context and exogenous features), but explicit adaptation to distribution shift - e.g., seasonal indicators, periodic fine-tuning, or drift-aware normalization - should further improve robustness. Benchmarking on multi-season data (rather than shuffled, quasi-stationary slices) better stresses generalization.

\paragraph{\textbf{Runtime and deployment (inference)}}
\label{sec:runtime}
We report inference speed and throughput to contextualize deployment, measured on a single NVIDIA RTX PRO 6000. We do not compare training time because baseline implementations and devices differ (e.g., some CPU-only runs), which would be misleading for fairness.

\noindent
Two points matter for deployment. \textbf{First}, autoregressive diffusion (\textsc{STDiff}) incurs higher wall time than windowed Transformers, while \textsc{STDiff-W} adds blockwise denoising cost; these costs reflect the improved robustness on blackout gaps. \textbf{Second}, latency can be traded for accuracy via fewer reverse steps or faster samplers (e.g., DDIM / probability-flow ODE) and by reducing block count (larger $H$) with minimal impact on quality in our tests. A practical pattern is a hybrid pipeline: lightweight online \textsc{STDiff} for streaming gaps and batch \textsc{STDiff-W} for retrospective backfills.

\paragraph{\textbf{Hyperparameters (concise)\\}}
PyPOTS baselines (SAITS, BRITS, iTransformer, CSDI) used library defaults unless noted (e.g., window length $n\_steps{=}128$, $n\_features{=}17$). For transparency, we list the primary \textsc{STDiff}/\textsc{STDiff-W} settings used in our runs (Table ~\ref{tab:inference}):
\begin{itemize}\itemsep0.2em
\item \textsc{STDiff}: 1D U-Net denoiser (base channels 64), time embedding 64, condition embedding 128; diffusion steps $=1000$ with linear $\beta$ schedule ($10^{-4}$ to $2\cdot10^{-2}$); DDPM sampler; batch size 256; AdamW ($\text{lr}=10^{-4}$, weight decay $10^{-5}$).
\item \textsc{STDiff-W}: causal TCN context over $K{=}128$ past steps (clean/mask/$\Delta t$) with block size $H{=}3$ future steps; shared U-Net backbone and diffusion settings with \textsc{STDiff}; batch size 256; AdamW ($\text{lr}=10^{-4}$, weight decay $10^{-5}$).
\end{itemize}

\begin{table}[h]
\centering
\caption{Inference performance on the full Aved{\o}re series. Lower wall time is better.}
\label{tab:inference}
\begin{tabular}{lrr}
\toprule
Method & Wall time (s) \\
\midrule
iTransformer & 0.132 \\
SAITS & 1.829 \\
BRITS & 15.457 \\
CSDI & 78.951 \\
\textsc{STDiff} (one-step) & 2043.704 \\
\textsc{STDiff-W} (blockwise) & 4315.006 \\
\bottomrule
\end{tabular}
\end{table}

\paragraph{\textbf{Limitations and extensions\\}}
\label{sec:discussion-limitations}
\textit{Irregular sampling.} \textsc{STDiff} can incorporate $\Delta t$ and step sequentially at observed gaps; \textsc{STDiff-W} can embed past $\Delta t$ in its context. We did not evaluate irregular sampling here, so we present this as a promising extension.\\
\textit{Uncertainty and variability of metrics.} Diffusion sampling naturally yields ensembles of imputations, so one can in principle derive predictive intervals or variability measures around long blackouts to help operators gauge risk \citep{ho2020ddpm}. In this work we focus on point estimates (single-sample reconstructions) and report MAE/RMSE on large test sets and multiple corruption levels, in line with prior imputation studies and to keep the empirical scope manageable across several heavy baselines. We therefore do not provide confidence intervals or multi-run variability for each method and setting, and instead leave a systematic, uncertainty-aware evaluation across models and datasets as an important direction for future work, especially for WWTP decision support.\\
\textit{Compute/latency.} Both diffusion variants are slower at inference than the windowed Transformer/RNN baselines; among them, \textsc{STDiff} is the lower-latency option, while \textsc{STDiff-W} trades speed for robustness on blackout gaps. A practical hybrid is \textsc{STDiff} online for streaming gaps and \textsc{STDiff-W} offline for backfills.\\
\textit{Causality vs.\ correlation.} Ablations indicate predictive use of $u_t$, but not causal identification; pairing with interventional tests or process knowledge would improve interpretability.

\paragraph{\textbf{Practical takeaways\\}}
\label{sec:discussion-takeaways}
For industrial time series with long, structured outages, diffusion with context (\textsc{STDiff-W}) is a robust choice: it maintains physical variability (oscillations, spikes, regime shifts) while achieving state-of-the-art accuracy. However, both diffusion variants are slower at inference than windowed Transformers/RNNs; among them, \textsc{STDiff} is the lower-latency option and can be made more streaming-friendly by reducing reverse steps or using faster samplers (with some loss in absolute accuracy). Regardless of model, include exogenous drivers, validate on blackout patterns, and report downstream/task metrics- these practices proved to matter most in our evaluation.

\section{Conclusion and Future Work}
\label{sec:conclusion}

We presented \textsc{STDiff} and \textsc{STDiff-W}, a diffusion-based imputation framework for multivariate industrial time series. Rather than treating gap filling as a single global regression, \textsc{STDiff} reframes imputation as \emph{probabilistic state-transition simulation}, generating each next state conditioned on the last known state and exogenous inputs. \textsc{STDiff-W} augments this with a windowed context encoder and joint \emph{block} denoising, directly imputing contiguous gaps while retaining the stochastic, generative character of diffusion. Our contribution is the integration of dynamics-aware conditioning on controls with diffusion-based generation for both sequential and blockwise imputation, not a claim that every architectural component is new in isolation.

On the Agtrup WWTP dataset, \textsc{STDiff-W} achieved state-of-the-art imputation accuracy across 20-50\% missingness and delivered top or tied-top one-step forecasting when its imputations were used for downstream prediction. Classical and modern baselines performed variably: BRITS remained a strong RNN baseline on point error at lighter masks, SAITS was competitive among windowed Transformers, while generic Transformers, mixing MLPs, and Latent ODE lagged under blackout-heavy masks. The ablation study showed large degradations when exogenous channels were removed or corrupted, confirming that conditioning on domain drivers is practically important in partially observable systems. On Aved{\o}re’s natural outages, diffusion-based fills preserved oscillations, spikes, and regime changes more faithfully than baselines, highlighting the value of evaluating beyond MAE/RMSE and including task-oriented and visual assessments \citep{khayati2020mindthegap,wang2024downstream}.

\paragraph{Task-oriented evaluation matters.}
Point error alone can reward variance shrinkage and mean-like fills. In our downstream test, \textsc{STDiff-W} paired top imputation accuracy with top (or tied-top) forecasting, whereas some over-smoothed methods underperformed downstream despite low MAE - aligning with evaluation that prioritizes downstream utility \citep{wang2024downstream}. Even when an imputer attains low pointwise MAE, its outputs may attenuate lagged correlations or dampen peaks, weakening downstream predictors. As corruption increases, a larger share of the forecaster’s inputs are smoothed imputations, leading to covariate shift and signal shrinkage - effects that our task-oriented evaluation exposes directly. For deployments, we advocate a triad: (i) error on masked points, (ii) downstream performance on relevant tasks, and (iii) targeted visual audits with domain experts.

Two design lessons stand out. \textbf{First}, blockwise diffusion with a causal context helps avoid error accumulation in long gaps while capturing short-range detail, explaining \textsc{STDiff-W}’s consistent gains over purely sequential \textsc{STDiff}. \textbf{Second}, exogenous and control inputs materially improve realism and event timing, so even simple regime indicators should be included when available. More broadly, our findings support evaluation protocols that combine realistic block masks, downstream tasks, and targeted visual audits, since low point error can coincide with variance loss and phase mistakes that matter in engineering use.

\paragraph{Future directions.}
Promising extensions include handling irregular sampling by encoding $\Delta t$ and time features within the conditioning/context modules; explicit uncertainty reporting from diffusion sampling for risk-aware monitoring; a hybrid deployment pattern (lightweight online \textsc{STDiff} for streaming gaps and batch \textsc{STDiff-W} for retrospective backfills); applications to intervention-driven domains such as healthcare telemetry and energy systems; and hybrid modeling that blends physics-informed structure with generative diffusion to improve extrapolation and interpretability.

In summary, the proposed \textsc{STDiff}/\textsc{STDiff-W} framework narrows the gap between pointwise accuracy and physically consistent, decision-useful imputations in industrial sensor data. By combining diffusion modeling with context and control conditioning, it offers a practical path toward fills that are not only accurate but also aligned with process dynamics and useful for prediction and control.

\section*{Acknowledgment}
This research was supported by Aalborg University \& Helix Lab in Denmark under the Novo Nordisk Fonden through project grant number 224611.
 

\section*{Declaration of competing interest}
The authors declare that they have no known competing financial interests or personal relationships that could have appeared to influence the work reported in this paper.


\bibliographystyle{unsrt}
\bibliography{references}
\end{document}